\documentclass[pdflatex,sn-basic,iicol,Numbered]{sn-jnl}

\usepackage{graphicx}%
\usepackage{multirow}%
\usepackage{amsmath,amssymb,amsfonts}%
\usepackage{amsthm}%
\usepackage{mathrsfs}%
\usepackage[title]{appendix}%
\usepackage{xcolor}%
\usepackage{textcomp}%
\usepackage{manyfoot}%
\usepackage{booktabs}%
\usepackage{algorithm}%
\usepackage{algorithmicx}%
\usepackage{algpseudocode}%
\usepackage{listings}%
\usepackage{subfig}
\usepackage{caption}
\usepackage{cite}

\theoremstyle{thmstyleone}%
\theoremstyle{thmstyletwo}%

\theoremstyle{thmstylethree}%

\begin{document}

\title[Article Title]{Localization Meets Uncertainty: Uncertainty-Aware Multi-Modal Localization}


\author[1]{\fnm{Hye-Min} \sur{Won}}\email{hyemin\_won@etri.re.kr}
\equalcont{These authors contributed equally to this work.}

\author[2]{\fnm{Jieun} \sur{Lee}}\email{jieun2@polaris3d.co}
\equalcont{These authors contributed equally to this work.}


\author*[1]{\fnm{Jiyong} \sur{Oh}}\email{jiyongoh@etri.re.kr}

\affil[1]{\orgdiv{Daegu-Gyeongbuk Research Division}, \orgname{Electronics and Telecommunications Research Institute (ETRI)}, \orgaddress{\state{Daegu}, \country{Repulic of Korea}}}

\affil[2]{\orgname{Polaris3D}, \orgaddress{\state{Pohang}, \country{Repulic of Korea}}}



\abstract{Reliable localization is critical for robot navigation in complex indoor environments.
In this paper, we propose an uncertainty-aware localization method that enhances the reliability of localization outputs without modifying the prediction model itself.
This study introduces a percentile-based rejection strategy that filters out unreliable 3-DoF pose predictions based on aleatoric and epistemic uncertainties the network estimates.
We apply this approach to a multi-modal end-to-end localization that fuses RGB images and 2D LiDAR data, and we evaluate it across three real-world datasets collected using a commercialized serving robot.
Experimental results show that applying stricter uncertainty thresholds consistently improves pose accuracy.
Specifically, the mean position error is reduced by 41.0\%, 56.7\%, and 69.4\%, and the mean orientation error by 55.6\%, 65.7\%, and 73.3\%, when applying 90\%, 80\%, and 70\% thresholds, respectively.
Furthermore, the rejection strategy effectively removes extreme outliers, resulting in better alignment with ground truth trajectories.
To the best of our knowledge, this is the first study to quantitatively demonstrate the benefits of percentile-based uncertainty rejection in multi-modal end-to-end localization tasks.
Our approach provides a practical means to enhance the reliability and accuracy of localization systems in real-world deployments.}




\maketitle

\section{Introduction}\label{sec:intro}

Localization is a classical problem in the literature of robotics.
With simultaneous localization and mapping (SLAM) \citep{thrun2005,Cadena2016} and its related technologies \citep{Lowry2016,zhang2024lidar}, localization techniques have been developed constantly, and the advance leads to some commercial services using mobile robots and self-driving cars.
Especially, service robots have been widely used in indoor environments such as hotels, hospitals, and restaurants for tasks like food delivery or guest assistance in recent years.  
In such dynamic environments, robots frequently encounter localization challenges such as occlusions, dynamic obstacles, or map drift.  
These issues can lead to localization failure, causing the robot to become lost or deliver items to incorrect locations.  
One of the most well-known failure scenarios is the \emph{kidnapped robot problem}, where a robot is unexpectedly displaced without any sensor trace, making recovery difficult for conventional SLAM-based approaches.
To ensure robust and uninterrupted operation, especially after initialization or recovery from failure, global localization—where a robot must estimate its pose from scratch without prior information—plays a critical role in these scenarios.
End-to-end localization based on deep neural network can be a promising solution to global localization.

PoseNet \citep{kendall2015posenet} is the first study on the deep learning-based end-to-end localization method.
It estimates the six-dimensional pose directly from sensor data (image) using neural networks in an end-to-end manner.
Since PoseNet, the following studies have been conducted based on image \citep{Wang2020}, 3D point cloud \citep{Wang2022, YU2022, Li2023, li2024diffloc}, inertial information \citep{Herath2022}, and the fusion of image and 2D point cloud \citep{Lee2023}.
These end-to-end localization methods are known to be more robust against sensor data variations such as noise and illumination. 
However, it is difficult to use them solely because they generally provide relatively higher localization errors compared to the matching-based localization methods.
\citet{Jo2020} utilized PoseNet to get an initial pose for a particle filter-based localization. 
However, they overlooked how much confidence we can have in the output of PoseNet.
If the PoseNet output with a high localization error is used as an initial pose, the navigation system may not be operated.

In this paper, we introduce an uncertainty-aware rejection mechanism to improve localization accuracy.
In recent, \citet{li2024diffloc} proposed measuring uncertainty from the outputs of their proposed method DiffLoc, which is a diffusion model for localization to estimate the 6D pose of a 3D LiDAR sensor from a point cloud.
Their experiments showed that the measured uncertainty is highly correlated with localization error.
We go one step further beyond \citep{li2024diffloc}.
We leverage uncertainty as a threshold. 
It allows the rejection of localization results with high errors, and we can trust the non-rejected localization results through the rejection strategy.
Experiments using our datasets collected by a 2D LiDAR and a camera demonstrate that the uncertainty-based rejection effectively reduces position and orientation errors.
More specifically, our uncertainty-based rejection method reduces the position error by up to 69.4\% and the orientation error by 73.3\%.
In particular, experimental results show that our strategy can exclude the outputs with significant position and orientation errors.
This means that the passed outputs are reliable enough for the results of global localization.
To the best of our knowledge, this is the first study that systematically utilizes uncertainty-based thresholds to reject unreliable localization results predicted in an end-to-end manner and improves the accuracy and confidence of pose estimates.

\section{Related Works}\label{sec:related_works}

\subsection{End-to-end localization}

End-to-end localization, also known as absolute pose regression, directly predicts the pose of a robot or a sensor from its data using deep neural networks without conventional procedures such as feature detection and matching.
It can serve as a solution for global localization, particularly in environments where GPS is unavailable. 
Depending on the type of sensors used for localization, it can be categorized into camera, LiDAR, and multi-modal localizations.

Visual localization estimates the current pose using only images captured by a camera in indoor or outdoor environments. 
Initially, convolutional neural networks (CNNs) are primarily leveraged to extract salient features.
However, recent studies have introduced various techniques in their models.
\citet{kendall2015posenet} developed a CNN-based 6-DoF pose regression model that allows localization without relying on feature matching or keyframes.
Meanwhile, \citet{Wang2020} employed the self-attention technique \citep{Vaswani2017} improve the accuracy of the end-to-end localization.
Moreover, Transformer architectures \citep{Vaswani2017} have been utilized for end-to-end camera localization as well \citep{Li2022, Qiao2023}.
In recent, \citet{wang2024efrnet} integrated CNNs, self-attention, and long short-term memory (LSTM) modules in a unified architecture to extract static features, which can lead to more effective 6-DoF pose estimation compared to using dynamic features.

LiDAR-based end-to-end localization leverages 3D structural information, making it robust in textureless environments. 
\citet{Wang2022} introduced the first LiDAR-based 6-DoF pose regression model, enhancing feature learning with self-attention mechanism. 
\citet{YU2022} proposed a deep neural network for pose regression that consists of two modules: a universal encoder for scene feature extraction and a regressor for pose estimation.
They also demonstrated the relationship between the regression capability and the number of hidden units in the regression module.
\citet{Yu2023} introduced additional classification headers alongside the original regression headers, together with a feature aggregation module based on temporal attention for spatial and temporal constraints.
\citet{Ibrahim2023} presented a self-supervised learning approach utilizing a Transformer-based backbone for LiDAR-based end-to-end localization.
Also, SGLoc \citep{Li2023} enhanced pose regression accuracy by incorporating scene geometry encoding.
Lastly, \citet{li2024diffloc} improved accuracy further by applying an iterative denoising process based on a diffusion model to the pose regression.

Some studies have combined complementary information from multiple modalities to improve the robustness of localization.
For instance, \citet{Lai2022} proposed leveraging both visual and LiDAR features to achieve more accurate and robust place recognition.
\citet{wang2022robot} developed a vision-assisted LiDAR localization method that effectively utilizes visual information to address issues related to 2D LiDAR-based localization drift.
Additionally, \citet{nakamura2024localization} incorporated a fisheye camera together with a 2D LiDAR system to enhance localization fault detection.
However, the methods mentioned above do not fall under the category of end-to-end localization techniques. 
FusionLoc \citep{Lee2023} is an end-to-end localization method that utilizes multi-modality.
In this study, we present a FusionLoc-based approach to make localization more reliable by rejecting network outputs with significant errors.

\subsection{Uncertainty quantification}

Uncertainty quantification is a well-established topic in pattern recognition and machine learning. 
While it did not receive much attention during the early stages of the deep learning revolution—especially in comparison to efforts to enhance the accuracy of deep learning algorithms—its importance is becoming increasingly recognized, particularly in safety-critical applications. 
A Bayesian approach is one of the most comprehensive frameworks for managing uncertainty.
However, developing and implementing a Bayesian deep neural network for regression tasks is very challenging because it is often impractical to determine posterior probabilities accurately.
Fortunately, \citet{Gal2016} presented dropout as an alternative to Bayesian approximation.
Additionally, \citet{kendall2017uncertainties} proposed using Monte Carlo (MC) dropout to quantify both aleatoric and epistemic uncertainties in regression tasks, such as pixel-wise depth estimation.

After the groundbreaking studies, the following researchers focused on network calibration, which aims to align estimated uncertainty values with empirical results. 
\citet{Kuleshov2018} introduced a simple, algorithm-agnostic method inspired by Platt scaling \citep{Platt1999}.
\citet{cui2020calibrated} utilized the maximum mean discrepancy, viewing calibration as a form of distribution matching.
Similarly, \citet{Bhatt2022} applied the $f$-divergence with the same perspective.
In another approach, \citet{Yu2021} proposed an auxiliary network branch to estimate uncertainty alongside the main branch used for the original regression task.
This method is similar to the work of \citep{Corbiere2022}, which also employs additional network branches to estimate uncertainty or confidence in classification problems.
For more details on uncertainty qualification in deep neural networks, refer to \citep{ABDAR2021, Gawlikowski2023}.

However, none of the studies mentioned above addressed the localization problem.
\citet{Chen2023} recently proposed a method for quantifying uncertainty in visual localization.
However, their approach estimates the pose of a query image through keypoint matching rather than an end-to-end method.
In contrast, \citet{li2024diffloc} suggested an end-to-end localization approach using a diffusion model.
However, their work primarily focused on the relationship between quantified uncertainties (variance) and positional errors, lacking qualitative experimental results.
Unlike \citep{Chen2023} and \citep{li2024diffloc}, this study aims to quantify uncertainty in the results of an end-to-end localization method.
We will also demonstrate how this quantification can effectively reject network outputs with significant errors, ultimately improving the localization performance.

\section{Method} \label{sec:method}

\subsection{FusionLoc}

FusionLoc \citep{Lee2023} is a deep learning-based robot localization method that combines an RGB image and 2D range data to improve the localization accuracy by leveraging the strengths of both sensors.
Specifically, FusionLoc predicts the robot's 3-DoF pose, including planar position and orientation, using an image $\mathbf{I}$ and 2D range data $\mathbf{S}$ as inputs as the following:
\begin{equation}
    [\hat{\mathbf{p}}, \hat{\mathbf{q}}] = \mathbf{f}( \mathbf{I}, \mathbf{S} ),
\nonumber
\end{equation}
where $\hat{\mathbf{p}} = [ \hat{x}, \hat{y} ]$ is the 2D coordinates of the robot position, and $\hat{\mathbf{q}} = [ \cos\hat{\theta}, \sin\hat{\theta} ]$ corresponds to the robot orientation.
The method computes an image feature from the input image using a feature extractor from AtLoc \citep{Wang2020}, while it calculates point features from the input range data using a different feature extractor from PointLoc \citep{Wang2022}.
To enhance the interaction between these two modalities, multi-head self-attention \citep{Vaswani2017} is employed.
This approach enables more effective multi-modality fusion than traditional methods such as concatenation or addition of the image and point features.
Lastly, the output of the multi-head self-attention block is passed through the regression block.
The regression block has two branches responsible for position and orientation.
Each branch consists of successive MLPs to provide the position $\hat{\mathbf{p}}$ and the orientation $\hat{\mathbf{q}}$.

\subsection{Measureing uncertainty}

Deep learning has shown remarkable performance on various complex tasks, primarily focused on enhancing predictive accuracy.
However, real-world applications often face uncertainty due to some factors, such as incomplete information and ambiguities.
This complexity makes it difficult to assess the performance of the model solely on the basis of accuracy \citep{cui2020calibrated}.
Therefore, quantifying uncertainty is crucial to improve prediction reliability, improve model robustness, and ensure safety.

Uncertainty can be divided into two categories: epistemic uncertainty and aleatoric uncertainty \citep{kendall2017uncertainties}.
Epistemic uncertainty arises from limitations in the model's knowledge or training process, typically due to insufficient data.
This uncertainty can be reduced by incorporating additional training data or enhancing the model architecture.
In contrast, aleatoric uncertainty stems from sensor noise, measurement errors, or inherent randomness in the data collection procedure.
Unlike epistemic uncertainty, aleatoric uncertainty cannot be eliminated through additional training, as it originates from data sensing.
Both types of uncertainty can be estimated using Bayesian neural networks (BNNs).
BNNs treat model weights as probabilistic distributions to quantify uncertainty.
However, computing the exact posterior distribution in high-dimensional spaces is almost impractical.
In this study, we use MC dropout \citep{kendall2017uncertainties} to approximate Bayesian inference and provide uncertainty estimation.

Let us consider a regression task with $N$ data pairs of input $\mathbf{x}$ and output $y$, i.e., $\{\left( \mathbf{x}_{i}, y_{i} \right)\}_{i=1}^{N}$.
To quantify aleatoric and epistemic uncertainties in this task, we consider a BNN model $\mathbf{f}$ to infer the posterior distribution.
From an input $\mathbf{x}$, it provides a model output $\hat{y}$ together with a variance $\hat{\sigma}^{2}$ of the aleatoric uncertainty.
In contrast, to estimate the epistemic uncertainty, we employ the MC dropout to approximate the posterior over the model.
By representing the model weights as $\hat{\mathbf{W}}$ from the approximate posterior, the model provides both the predictive mean and the variance, i.e., $[ \hat{y}, \hat{\sigma}^{2} ] = \mathbf{f}^{\hat{\mathbf{W}}} (\mathbf{x})$.
The objective function of learning the model can be defined without the regularization term as the following:
\begin{equation}
\begin{split}
    & \frac{1}{N} \sum_{i=1}^{N} \left[ \frac{1}{2 \hat{\sigma}_{i}^{2}} \left\Vert y_{i} - \hat{y}_{i} \right\Vert^{2} + \frac{1}{2} \log \hat{\sigma}_{i}^{2} \right] \\
    & = \frac{1}{2N} \sum_{i=1}^{N} \left[ \exp (-s_{i}) \left\Vert y_{i} - \hat{y}_{i} \right\Vert^{2} +  s_{i} \right],
\end{split}
\nonumber
\end{equation}
where $s_{i}=\log \hat{\sigma}_{i}^{2}$.
Here, the second equation is more numerically stable for the case of division by zero.
After training, we can estimate the uncertainty of an output $\hat{y}$ by $T$ times multiple inferences for a given input as the following:
\begin{equation}
    \frac{1}{T}\sum_{t=1}^{T} \left[ \hat{y}_{t}^{2} - \left( \frac{1}{T} \sum_{t=1}^{T} \hat{y}_{t} \right)^{2} \right] + \frac{1}{T} \sum_{t=1}^{T} \hat{\sigma}_{t}^{2},
\nonumber
\end{equation}
where $(\hat{y}_{t}, \hat{\sigma}_{t}^{2})$ is the $t$-th outputs of the model based on randomly determined weights by dropout.
Note that the first and the second terms correspond to the epistemic and the aleatoric uncertainties of the output, respectively. More details are referred to as \citep{kendall2017uncertainties}.

\subsection{Uncertainty-aware localization}

\begin{figure*}[thpb]
    \centering
      \includegraphics[width=6.0in]{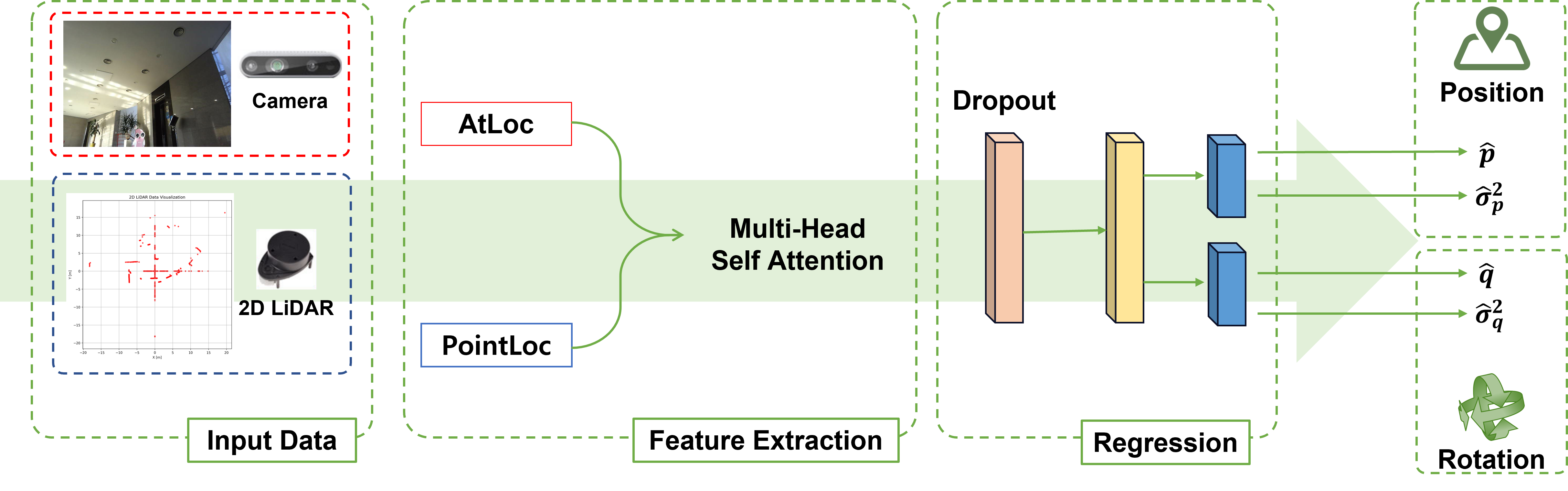}
      \caption{Our pipeline for uncertainty-aware end-to-end localization}
      \label{fig:method}
\end{figure*}

In this section, we describe our uncertainty-aware localization method based on the fusion of RGB image and 2D range data captured from a commercialized serving robot.

To perform the uncertainty-aware localization, we modify the FusionLoc \citep{Lee2023} architecture such that it has two more output nodes $\hat{\sigma}_{\mathbf{p}}^{2}$ and $\hat{\sigma}_{\mathbf{q}}^{2}$ to measure the aleatoric uncertainty.
Thus, our model can be represented as the following:
\begin{equation}
    [\hat{\mathbf{p}}, \hat{\sigma}_{\mathbf{p}}^{2}, \hat{\mathbf{q}}, \hat{\sigma}_{\mathbf{q}}^{2} ] = \mathbf{f}^{\hat{\mathbf{W}}}(\mathbf{I}, \mathbf{S}).
\nonumber
\end{equation}
Fig. \ref{fig:method} shows the pipeline of our model.
As mentioned above, we replace $\hat{\sigma}_{\mathbf{p}}^{2}$ and $\hat{\sigma}_{\mathbf{q}}^{2}$ by $s_{\mathbf{p}} = \log \hat{\sigma}_{\mathbf{p}}^{2}$ and $s_{\mathbf{q}} = \log \hat{\sigma}_{\mathbf{q}}^{2}$ for computational stability.
Also, we measure the epistemic uncertainty by applying the MC dropout to the output of the self-attention block mentioned above. 
Given $N$ training samples $\{(\mathbf{I}_{i}, \mathbf{S}_{i}, \mathbf{p}_{i}, \mathbf{q}_{i})\}_{i=1}^{N}$, the loss function can be defined as the following:
\begin{equation}
\begin{split}
    & \frac{1}{2N} \sum_{i=1}^{N} \left[ \exp (-s_{\mathbf{p}i}) \left\Vert \mathbf{p}_{i} - \hat{\mathbf{p}}_{i} \right\Vert^{2} +  s_{\mathbf{p}i} \right] \\
    & + \frac{1}{2N} \sum_{i=1}^{N} \left[ \exp (-s_{\mathbf{q}i}) \left\Vert \mathbf{q}_{i} - \hat{\mathbf{q}}_{i} \right\Vert^{2} +  s_{\mathbf{q}i} \right].
\end{split}
\nonumber
\end{equation}
After finishing training process, we can predict the position $\mathbf{p}^{*}$ and orientation $\mathbf{q}^{*}$ of the robot by performing the inference $T$ times using a pair of $(\mathbf{I}, \mathbf{S})$ as the following:
\begin{equation}
    \mathbf{p}^{*} = \frac{1}{T} \sum_{t=1}^{T} \hat{\mathbf{p}}_{t}, \quad \mathbf{q}^{*} = \frac{1}{T} \sum_{t=1}^{T} \hat{\mathbf{q}}_{t},
\nonumber
\end{equation}
where $\hat{\mathbf{p}}_{t}$ and $\hat{\mathbf{q}}_{t}$ are the $t$-th position and orientation outputs obtained using the trained model.
And, their corresponding uncertainties $u_{\mathbf{p}}$ and $u_{\mathbf{q}}$ are computed as the following:
\begin{equation}
\begin{split}
    u_{\mathbf{p}} &= \frac{1}{T} \sum_{t=1}^{T} \left[ \hat{\mathbf{p}}_{t}^{2} - (\mathbf{p}^{*})^{2} \right] + \frac{1}{T} \sum_{t=1}^{T} \hat{\sigma}_{\mathbf{p}t}^{2}, \\
    u_{\mathbf{q}} &= \frac{1}{T} \sum_{t=1}^{T} \left[ \hat{\mathbf{q}}_{t}^{2} - (\mathbf{q}^{*})^{2} \right] + \frac{1}{T} \sum_{t=1}^{T} \hat{\sigma}_{\mathbf{q}t}^{2},
\end{split}
\nonumber
\end{equation}
where $\hat{\sigma}_{\mathbf{p}t}^{2}$ and $\hat{\sigma}_{\mathbf{q}t}^{2}$ are the outputs corresponding to the aleatoric uncertainty measurement of position and orientation, respectively.

Note that we cannot expect how much error the network output has, which may often lead to a serious problem in safety.
Under the assumption that a network output with high uncertainty has a large localization error, we can select reliable outputs based on the uncertainty values.
Consequently, this rejection strategy enhances localization accuracy.
By discarding results with high uncertainty, we ensure that the remaining outputs have comparatively lower errors.

Our approach utilizes a percentile-based thresholding method that rejects a portion of the network outputs that exceed predefined uncertainty thresholds.
In this situation, it is crucial to determine the appropriate threshold value. 
We experiment with different percentile thresholds (100\%, 90\%, 80\%, and 70\%), progressively filtering the top 0\%, 10\%, 20\%, and 30\% of the most uncertain predictions.
Although this thresholding approach is straightforward, it effectively rejects the network results corresponding to outliers, leading to a more reliable and precise localization system.

In the next section, we will present a detailed analysis of how the uncertainty-based rejection strategy improves localization performance.

\begin{figure}[t]
    \centering
    \subfloat[Rear view]{
        \includegraphics[width=1.33in]{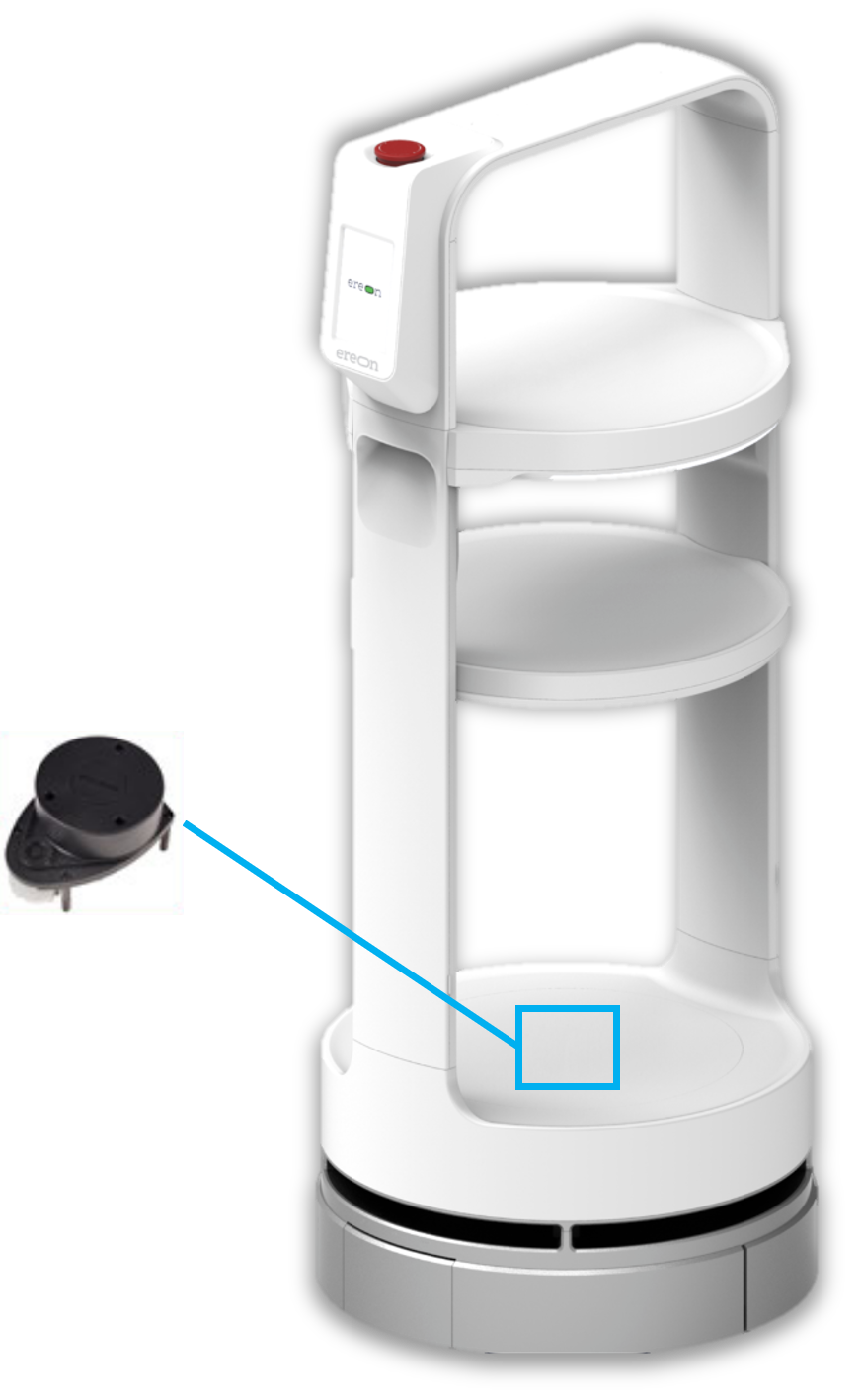}
        \label{ereon_back}
    }%
    \hfill
    \subfloat[Front view]{
        \includegraphics[width=1.4in]{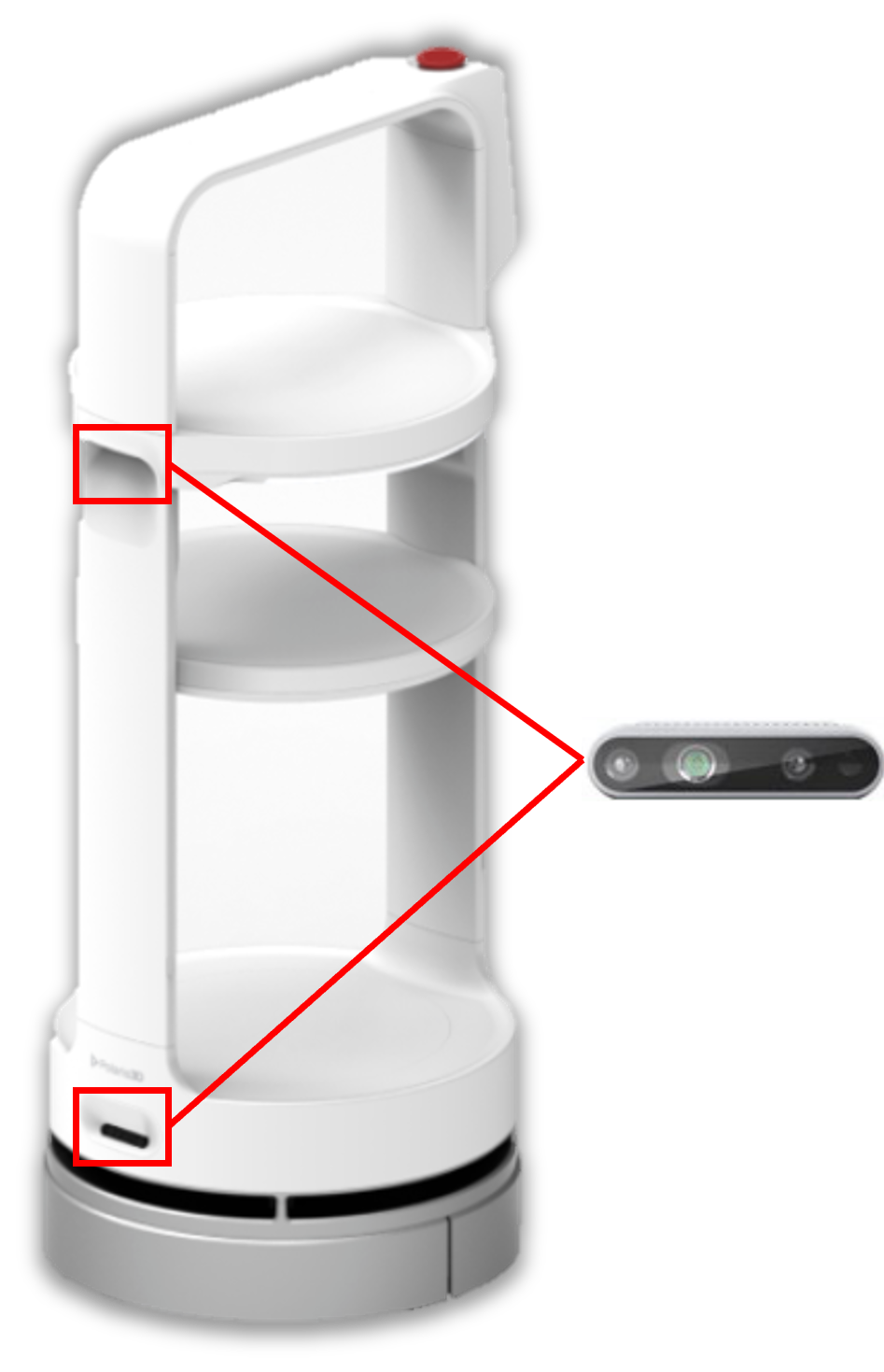}
        \label{ereon_front}
    }%
    \caption{A serving robot used in this study: The blue box represents a SLAMTEC RPLiDAR A1M8, while the red boxes indicate Intel RealSense D435 cameras.}
    \label{fig:ereon}
\end{figure}

\begin{figure*}[h]
    \centering
    \subfloat[TheGardenParty]{
        \includegraphics[width=2.0in]{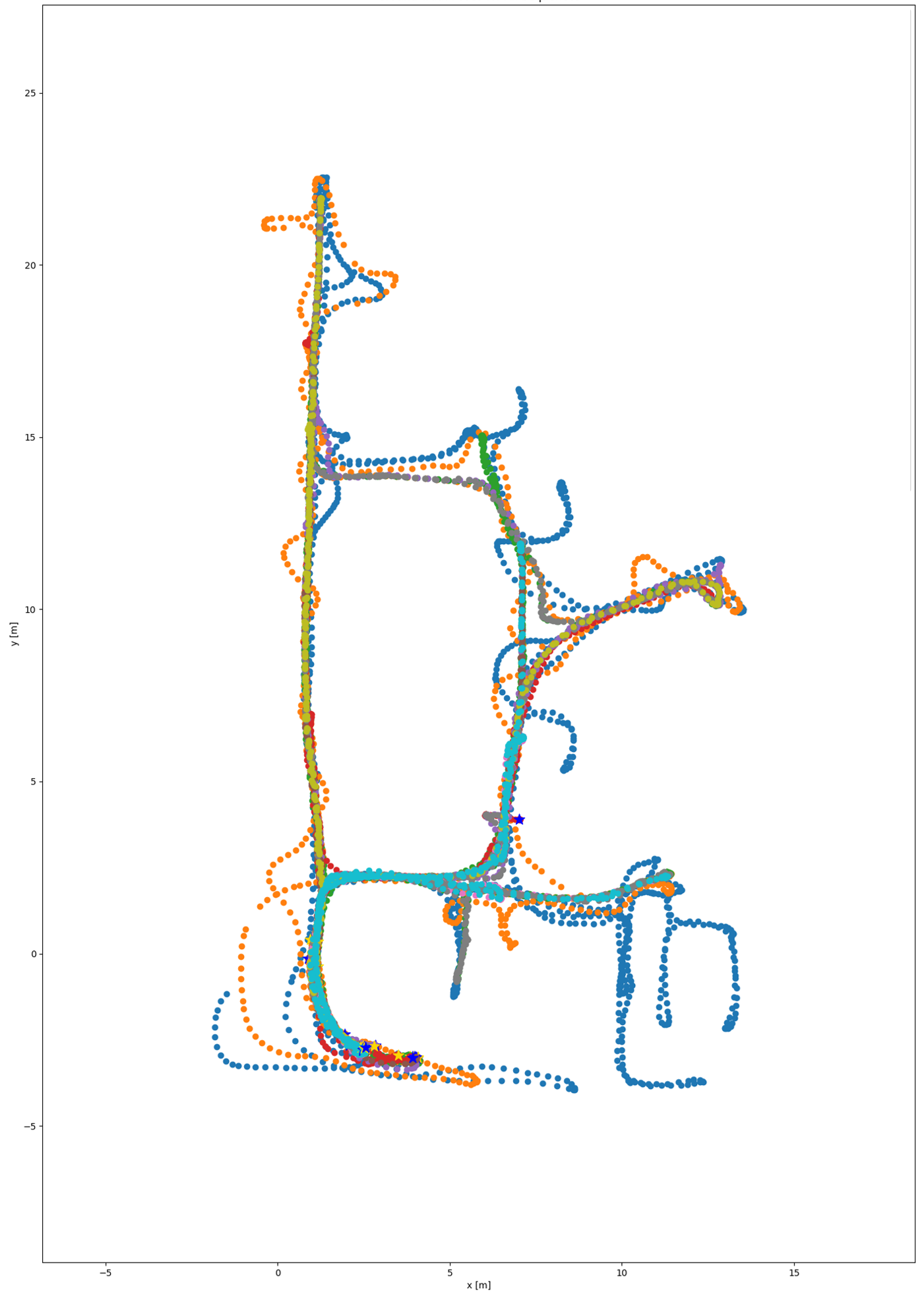}
        \label{fig:gardenparty}
    }
    \hfill
    \subfloat[ETRI]{
        \includegraphics[width=1.93in]{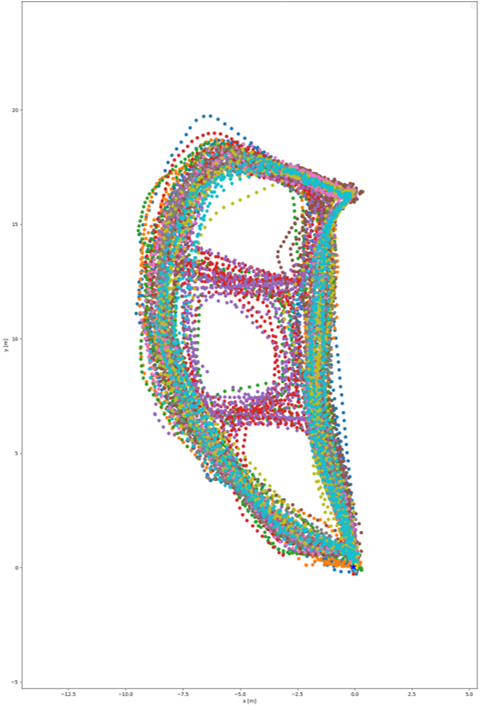}
        \label{fig:etri}
    }
    \hfill    
    \subfloat[SusungHotel]{
        \includegraphics[width=0.99in]{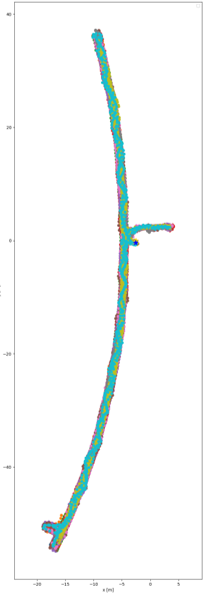}
        \label{fig:etri}
    }
    \caption{Robot trajectories in each dataset.}
    \label{fig:trajectories}
\end{figure*}

\begin{figure*}[h]
    \centering
    \subfloat[]{
        \includegraphics[width=1.40in]{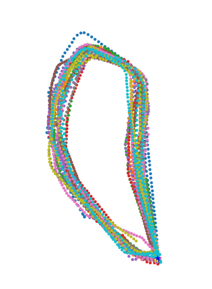}
        \label{fig:traj1}
    }
    \hfill
    \subfloat[]{
        \includegraphics[width=1.40in]{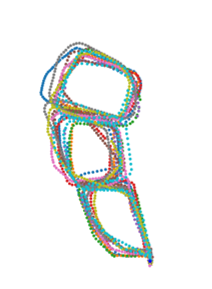}
        \label{fig:traj2}
    }
    \hfill
    \subfloat[]{
        \includegraphics[width=1.40in]{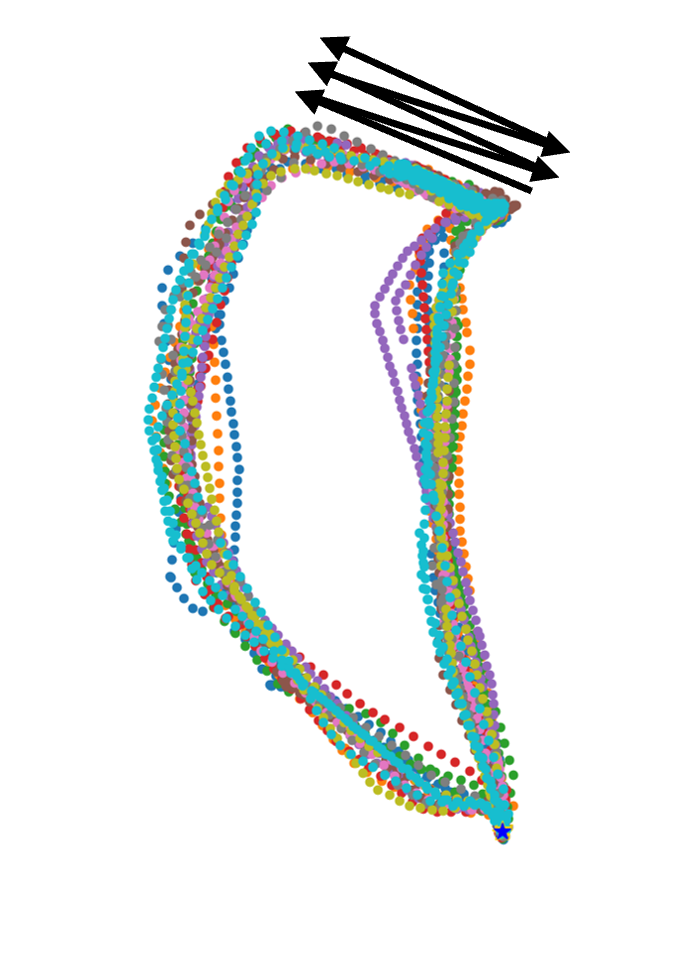}
        \label{fig:traj3}
    }
    \hfill
    \subfloat[]{
        \includegraphics[width=1.40in]{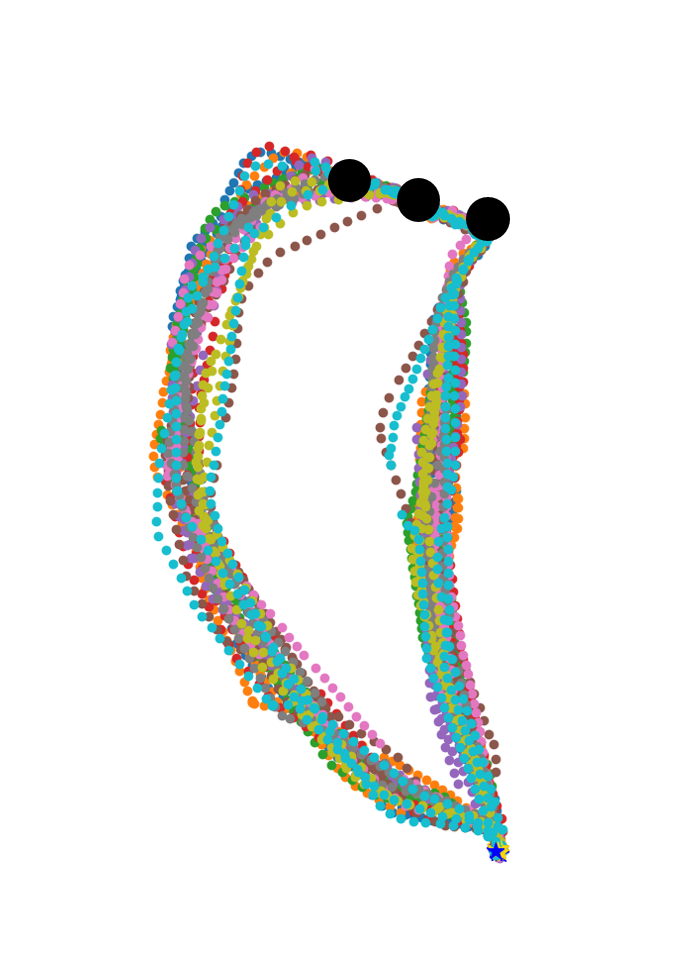}
        \label{fig:traj4}
    }
    \caption{Visualization of robot trajectories in different scenarios. (a) Full-loop trajectory. (b) Zigzag navigation. (c) Localized back-and-forth motion. (d) In-place rotations at specific locations.}
    \label{fig:trajectory_visualization} 
\end{figure*}

\section{Experiments} \label{sec:experiments}

\subsection{Datasets}

For our experiments, we constructed four datasets from indoor environments named TheGardenParty, ETRI and SusungHotel.
For multi-modality, sensor data such as RGB images and 2D range data were collected using a commercialized serving robot, Polaris3D Ereon, as shown in Fig. \ref{fig:ereon}.
This robot is equipped with two cameras and a 2D LiDAR sensor.
For our purposes, we utilized a lower-mounted camera to capture RGB images and the LiDAR sensor to gather 2D range data.
We utilized an Intel RealSense D435 camera for collecting the TheGardenParty dataset and an Astra Stereo SU3 camera for the ETRI dataset.
The LiDAR sensor is the SLAMTEC RPLiDAR A1M8, which operates at 8 Hz with a maximum range of 12 meters and an angular resolution of 0.313$^{\circ}$.
It performs 360$^{\circ}$ scans, generating up to 1,150 2D points per scan.

The collected datasets consist of RGB images that provide visual context, 2D LiDAR scans that capture structural and geometric information, and 3D poses derived from these scans.
We aimed to synchronize the images, 2D range data and poses as closely as possible in time. 
These datasets were utilized for training and evaluating deep learning models focused on robot localization.

\begin{table*}[t]
    \caption{Summary of each dataset's characteristics}
    \label{tab:dataset_comparison}
    \centering
    \renewcommand{\arraystretch}{1.2}
    \begin{tabular}{|c|c|c|c|}
    \hline
    Attribute & TheGardenParty & ETRI & SusungHotel \\
    \hline
    Image Resolution (pixels) & 320$\times$240 & 640$\times$480 & 640$\times$480 \\
    Navigation Pattern & Predefined & 4 patterns & Predefined \\
    Environment Type & Structured & Complex & Structured\\
    \hline
    \# Training tuples & 7,848 & 12,688 & 7,625 \\
    \# Validation tuples & 2,294 & 2,964 & 1,258 \\
    \# Test tuples & 3,184 & 2,794 & 1,276 \\
    Total tuples & 13,326 & 19,014 & 9,625 \\
    \hline
    \end{tabular}
\end{table*}

Fig. \ref{fig:trajectories} presents ground truth trajectories collected from three different datasets.
In the figure, each color represents a different sequence, while the robot's start and end positions are marked with gold and a downward star, respectively.
Table \ref{tab:dataset_comparison} compares the key characteristics of the three datasets and summarizes their features, including the number of samples used for training, evaluation, and validation. 
The TheGardenParty dataset provides images at a resolution of 320$\times$240 pixels, and it depicts a structured indoor environment with predefined paths.
The dataset comprises 13,326 data tuples across 35 sequences, with 24 sequences used for training, 6 for evaluation, and 5 for validation.
The ETRI dataset supports 640$\times$480 pixels and it represents a more complex navigation environment where the robot explores various paths including exploration and obstacle avoidance.
As shown in Fig. 4, the ETRI dataset features four distinct movement patterns: 
\begin{itemize}
    \item Straight corridor navigation: In this pattern, the robot navigates the entire space in a continuous loop before returning to its starting point.
    \item Zigzag movement: Here, the robot moves in a zigzag manner, weaving between obstacles.
    \item Repetitive back-and-forth motion: This pattern involves the robot moving back and forth within a confined space before proceeding with further exploration.
    \item Rotational maneuvers: In this last pattern, the robot performs in-place rotations at specific locations before retracing the same trajectory as in the first pattern.
\end{itemize}
The dataset comprises 19,014 tuples and 100 sequences, with 66 sequences designated for training, 16 for evaluation, and 16 for validation.

The SusungHotel dataset provides high-resolution images with a resolution of 640×480 pixels. 
It consists of a total of 9,625 data tuples, collected from 20 distinct sequences. 
These sequences are categorized into 16 for training, 2 for validation, and 2 for evaluation. 
Notably, among the three datasets, the SusungHotel dataset features the longest continuous trajectories captured in a single recording session.

\begin{figure*}[h]
    \centering
    \subfloat[]{
        \includegraphics[width=.3\textwidth]{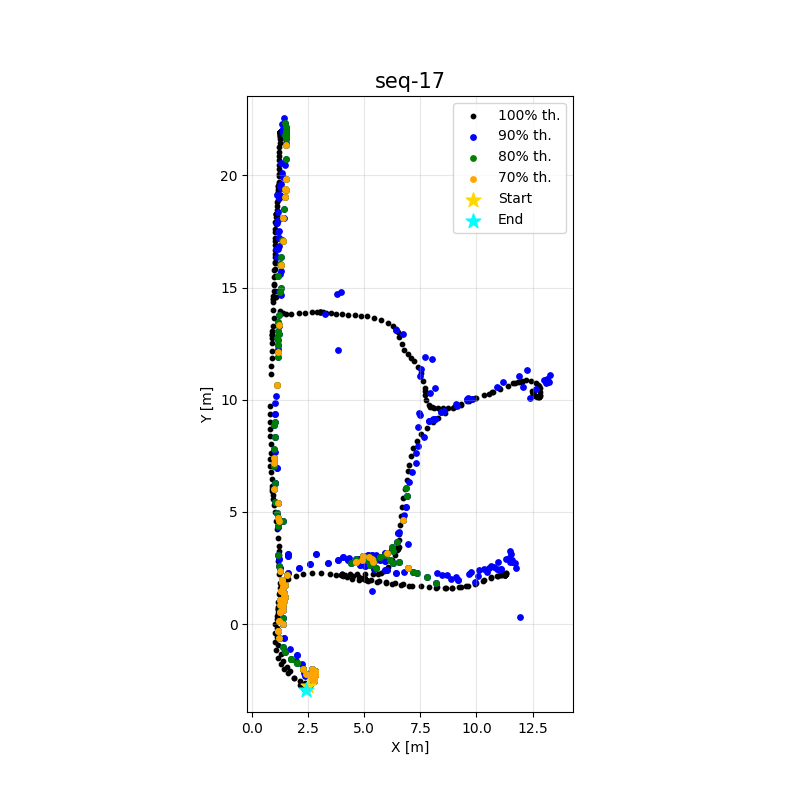}
        \label{fig:traj1}
    }
    \hfill
    \subfloat[]{
        \includegraphics[width=.3\textwidth]{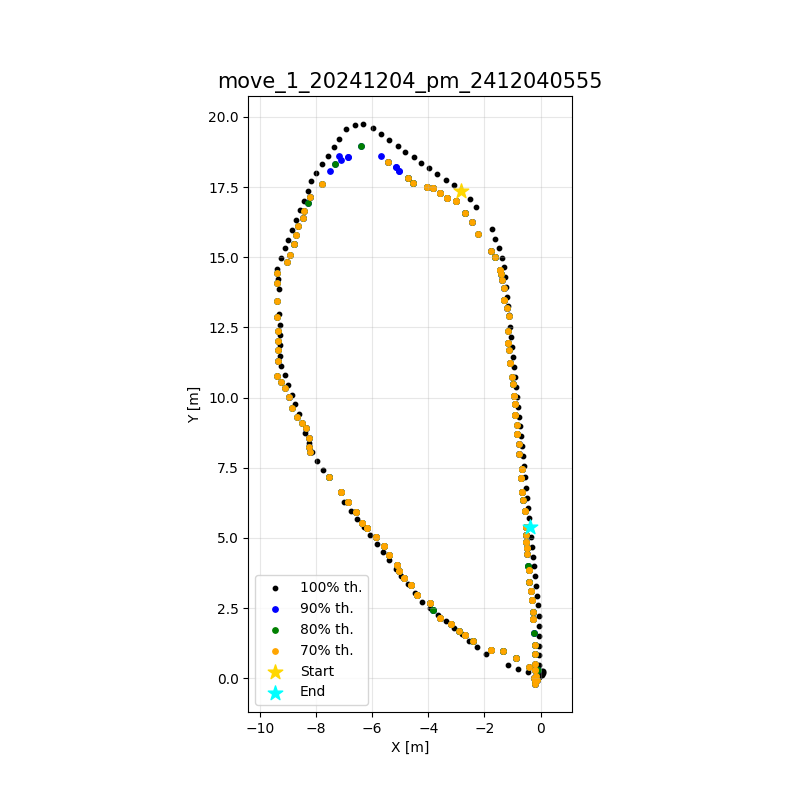}
        \label{fig:traj2}
    }
    \hfill
    \subfloat[]{
        \includegraphics[width=.3\textwidth]{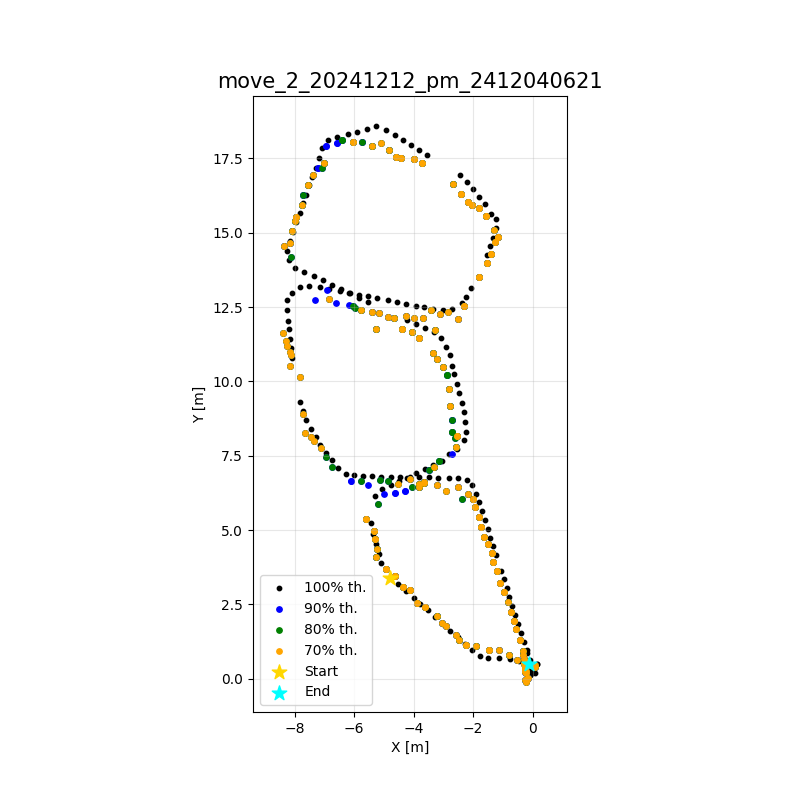}
        \label{fig:traj3}
    }
    \hfill
    \subfloat[]{
        \includegraphics[width=.3\textwidth]{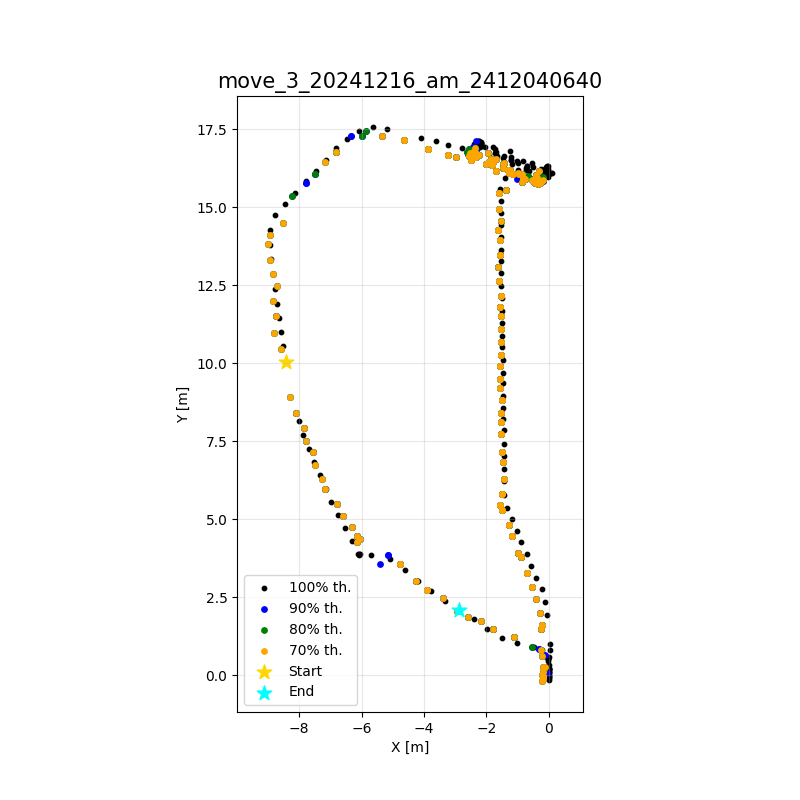}
        \label{fig:traj4}
    }
    \hfill
    \subfloat[]{
        \includegraphics[width=.3\textwidth]{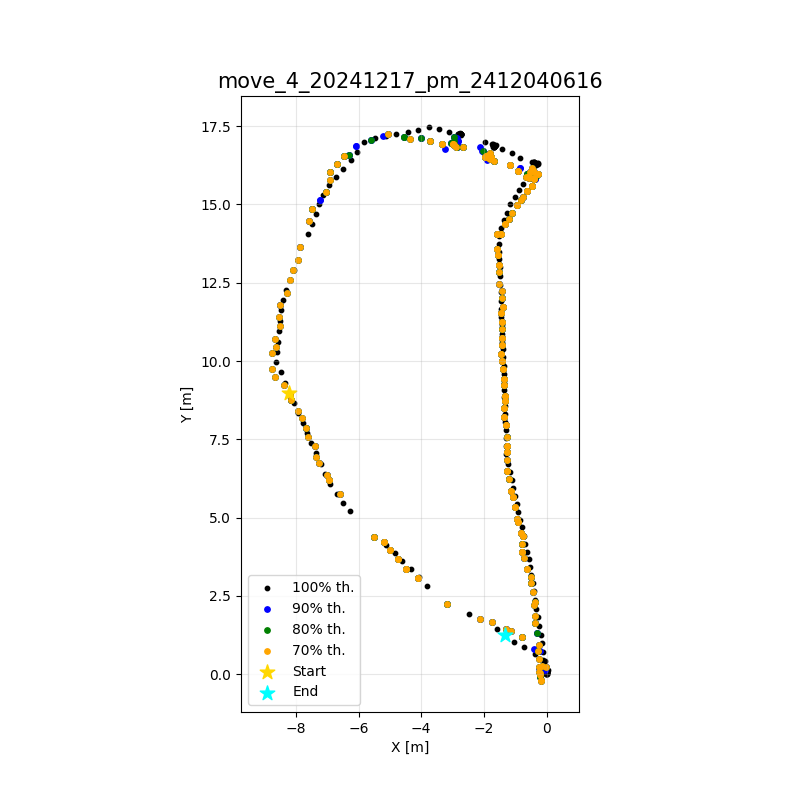}
        \label{fig:traj5}
    }
    \hfill
    \subfloat[]{
        \includegraphics[width=.3\textwidth]{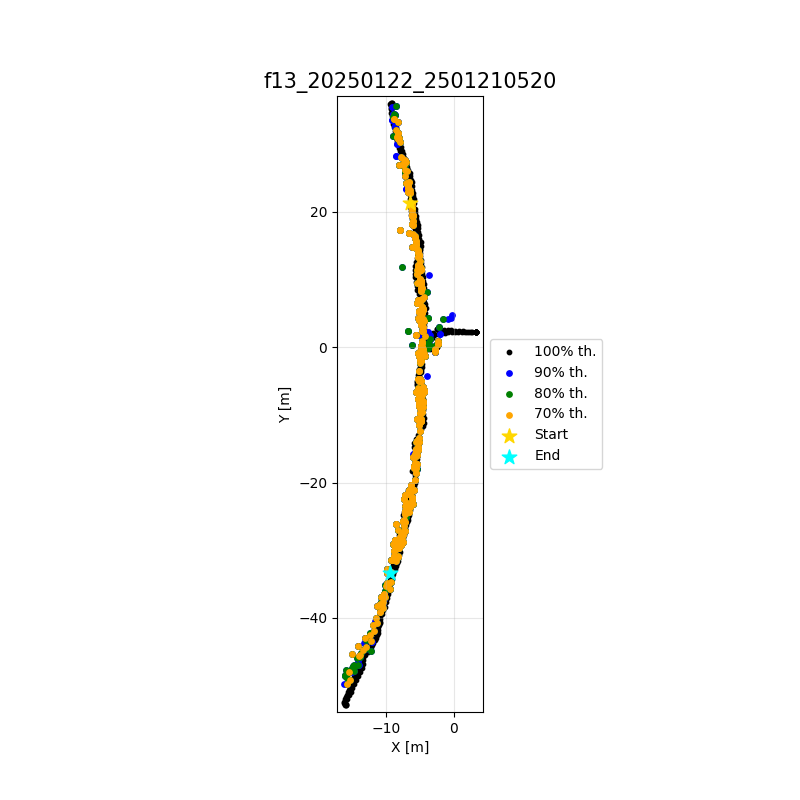}
        \label{fig:traj6}
    }
    \caption{Trajectory visualization with different uncertainty thresholds (100\%, 90\%, 80\%, 70\%). (a) TheGardenParty. (b) ETRI: Full-loop trajectory. (c) ETRI: Zigzag navigation. (d) ETRI: Localized back-and-forth motion. (e) ETRI: In-place rotations at specific locations. (f) SusungHotel.}
    \label{fig:trajectory_filtering_examples} 
\end{figure*}

\begin{table*}[h]
    \centering
    \renewcommand{\arraystretch}{1.2}
    \resizebox{\textwidth}{!}{
    \begin{tabular}{|c|c||c|c|c|c||c|c|c|c||c|c|c|c|}
        \hline
        \multicolumn{2}{|c||}{Dataset} 
        & \multicolumn{4}{c||}{TheGardenParty} 
        & \multicolumn{4}{c||}{ETRI}
        & \multicolumn{4}{c|}{SusungHotel} \\
        \hline
        \multicolumn{2}{|c||}{Threshold} 
        & 100\% th. & 90\% th. & 80\% th. & 70\% th. 
        & 100\% th. & 90\% th. & 80\% th. & 70\% th. 
        & 100\% th. & 90\% th. & 80\% th. & 70\% th. \\
        \hline
        \multirow{4}{*}{Position (m)} 
        & Min    & 0.072 & 0.074 & 0.069 & 0.098 
        & 0.023 & 0.025 & 0.026 & 0.031
        & 0.022 & 0.028 & 0.031 & 0.019 \\
        & Median & 0.573 & 0.567 & 0.552 & 0.523 
        & 0.289 & 0.298 & 0.300 & 0.298
        & 0.659 & 0.561 & 0.477 & 0.424 \\
        & Max    & 5.559 & 5.288 & 4.317 & 4.192 
        & 1.610 & 1.613 & 1.287 & 1.490
        & 66.392 & 66.279 & 59.829 & 59.753 \\
        & \textbf{Mean} & \textbf{0.776} & \textbf{0.738} & \textbf{0.712} & \textbf{0.682} 
        & \textbf{0.324} & \textbf{0.326} & \textbf{0.327} & \textbf{0.327} 
        & \textbf{4.262} & \textbf{2.788} & \textbf{2.211} & \textbf{1.575} \\
        \hline
        \multirow{4}{*}{Orientation ($^{\circ}$)} 
        & Min    & 0.0204 & 0.017 & 0.003 & 0.015 
        & 0.013 & 0.012 & 0.008 & 0.010 
        & 0.004 & 0.008 & 0.009 & 0.008 \\
        & Median & 2.672 & 2.544 & 2.512 & 2.299 
        & 1.127 & 1.052 & 0.978 & 0.902 
        & 2.732 & 2.466 & 2.168 & 2.020 \\
        & Max    & 72.738 & 54.508 & 49.384 & 48.397 
        & 26.224 & 16.182 & 15.569 & 11.024
        & 179.385 & 174.738 & 173.323 & 127.828 \\
        & \textbf{Mean} & \textbf{4.880} & \textbf{3.783} & \textbf{3.358} & \textbf{3.126} 
        & \textbf{1.990} & \textbf{1.66086} & \textbf{1.52688} & \textbf{1.409} 
        & \textbf{13.711} & \textbf{9.008} & \textbf{5.861} & \textbf{4.329} \\
        \hline
    \end{tabular}
    }
    \caption{Comparison of position and orientation metrics under different uncertainty thresholds for TheGardenParty, ETRI, and SusungHotel datasets.}
    \label{tab:uncertainty_combined}
\end{table*}

\subsection{Evaluation}

In this subsection, we evaluate the performance of our localization method by applying an uncertainty-based rejection approach.
To achieve this, we utilized the model mentioned in Sec. \ref{sec:method} and measured the epistemic and aleatoric uncertainties in the model’s predictions for position and orientation.
We demonstrate the improvement in the reliability of position and orientation predictions by applying percentile-based thresholds for uncertainty values and discarding outputs that exceed these thresholds. 
Specifically, experiments were conducted using 100\%, 90\%, 80\%, and 70\% as thresholds, progressively rejecting the top 0\%, 10\%, 20\%, and 30\% of the outputs with the highest uncertainty.
This rejection method retains only the reliable results, minimizing the influence of extreme outliers with high uncertainty.
As a result, this approach finally leads to a more robust evaluation of the model’s performance.

To evaluate the impact of the rejection approach on localization performance, we compared the median and mean errors in position and orientation before and after applying the rejection based on uncertainty thresholding.
Additionally, we measured the processing time at both batch and sequence levels to analyze the computational cost.
Our approach shows that rejecting high-uncertainty outputs reduces their negative impact on the results with significant errors, thereby improving reliability and accuracy.

\begin{figure*}[htbp]
    \centering
    \subfloat[TheGardenParty]{
        \includegraphics[width=5.5in]{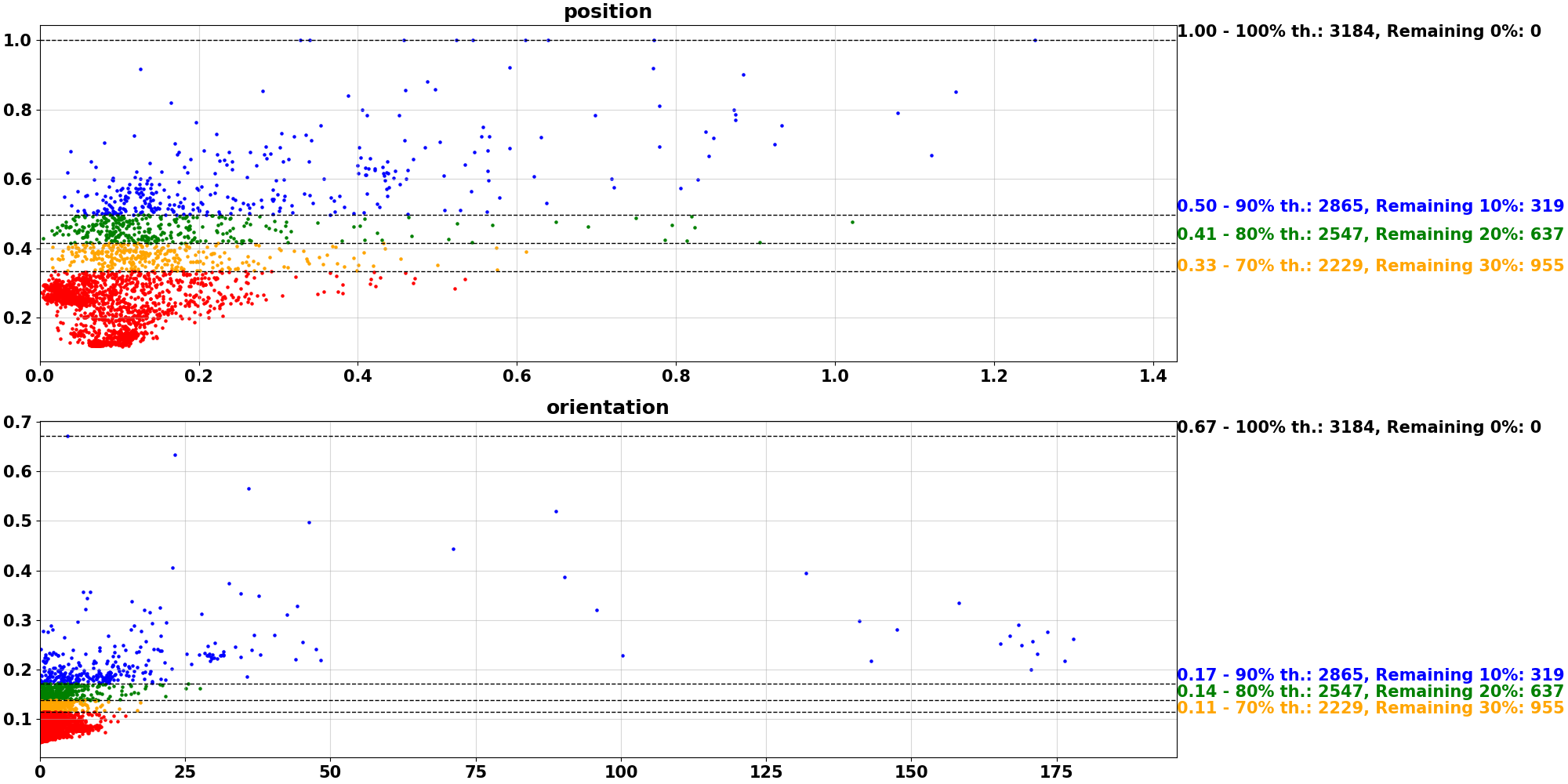}
    } \\
    \subfloat[ETRI]{
        \includegraphics[width=5.5in]{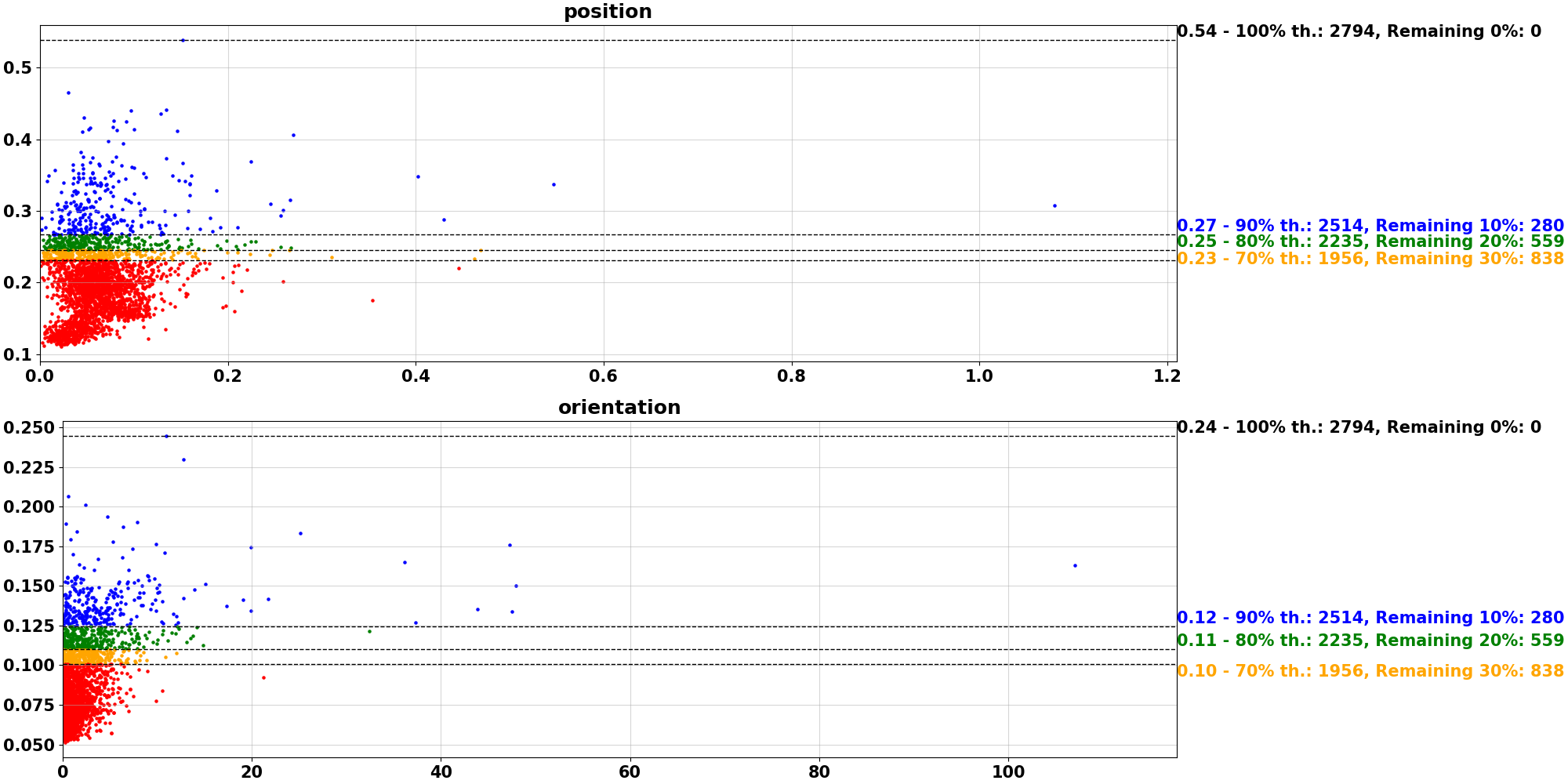}
    } \\
    \subfloat[SusungHotel]{
        \includegraphics[width=5.5in]{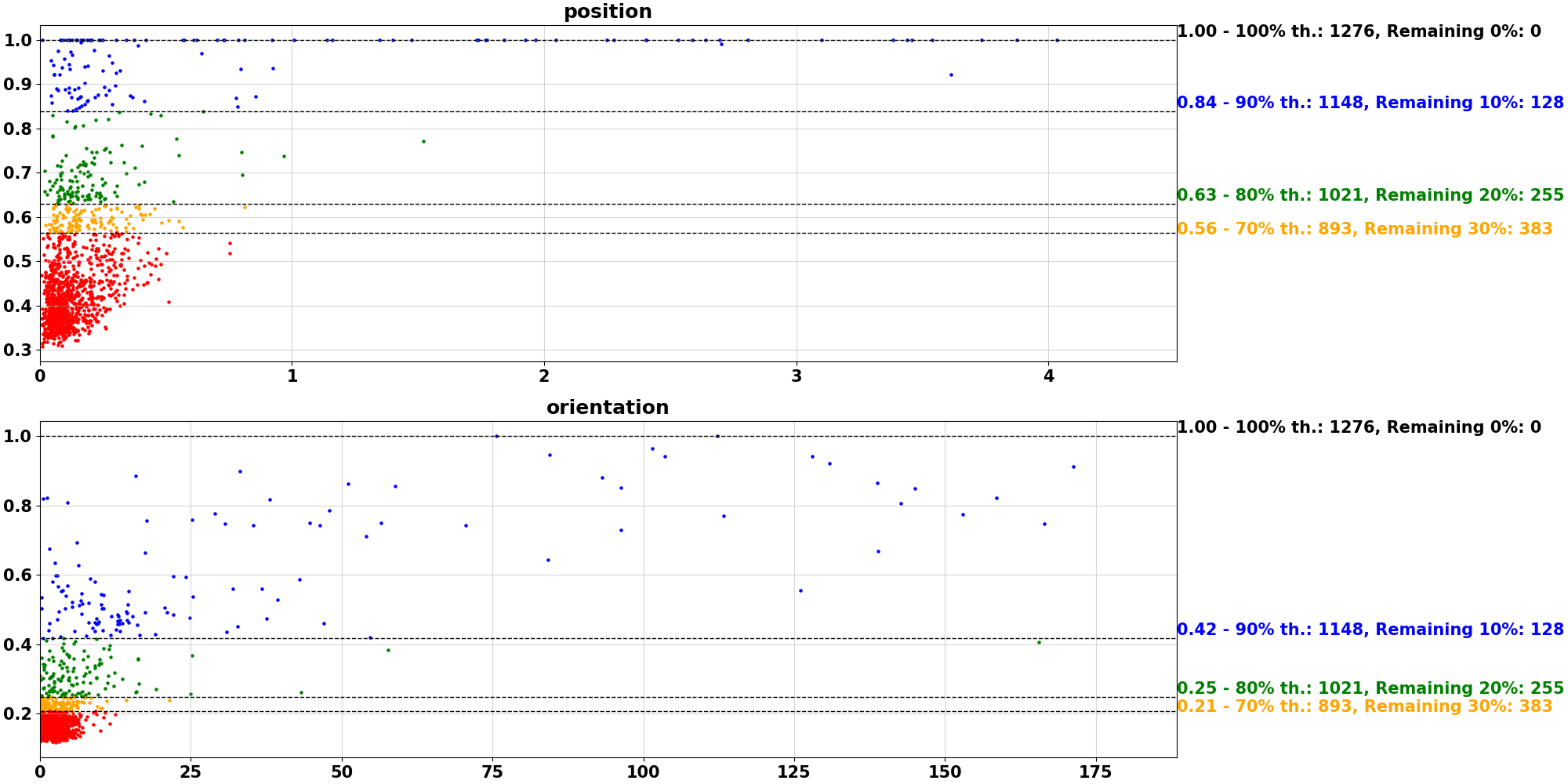}
    }
    
    \caption{Comparison of uncertainty-based rejection results with varying thresholds. The upper and lower scatter plots of each dataset represent the uncertainty for position and orientation errors, respectively.}
    \label{fig:uncertainty_filtering_combined_1}
\end{figure*}

Fig. \ref{fig:uncertainty_filtering_combined_1} illustrates the distribution of position and orientation errors after applying the uncertainty-based rejection approach.
Each scatter plot shows error value on the x-axis and their corresponding uncertainty on the y-axis.
The outputs from the network are color-coded to distinguish between low-uncertainty (more reliable) and high-uncertainty (potentially erroneous) outputs.
In the figure, red dots represent outputs with low uncertainty (below the 70\% threshold), which are considered reliable predictions.
They remain after applying all thresholds (100\%, 90\%, 80\%, and 70\%).
On the other hand, blue dots indicate high-uncertainty outputs that are rejected when the 90\% threshold is applied, meaning a higher likelihood of being erroneous.
Additionally, green and orange dots represent outputs with moderate uncertainty, positioned between the red and blue dots.
The black dashed lines in each plot indicate the rejection thresholds applied at different percents.
As the threshold decreases (from 100\% to 70\%), the number of low-uncertainty outputs (red) increases, while high-uncertainty outputs (blue) are progressively discarded.
On the right side of each plot, the first number represents the uncertainty threshold applied at that level, while the percentages and corresponding values indicate the number of the remaining outputs.
Overall, we can see that this strategy can effectively enhance the reliability of the network outputs by rejecting those with high uncertainty.
In Fig. \ref{fig:uncertainty_filtering_combined_1}, we also observe that applying a 70\% threshold led to the removal of 955 outputs from the TheGardenParty dataset, 838 outputs from the ETRI dataset and 383 outputs from the SusungHotel dataset.
This indicates that excessive rejection can result in data loss and require multiple inferences to provide a non-rejected output while the rejection approach effectively reduces errors on average.
Thus, it is crucial to determine an appropriate threshold.
Experimental results indicate that the 70\% threshold achieves a satisfactory rejection while maintaining a sufficient number of network outputs.
However, in a specific application, the rejection ratio should be carefully adjusted to balance performance improvement and multiple inferences.
Therefore, selecting an optimal threshold is essential to maximizing performance while preserving sufficient data for model learning.
To further illustrate the impact of uncertainty-based thresholding, we visualize the predicted trajectories with and without filtering in various test sequences.
Fig.~\ref{fig:trajectory_filtering_examples} show the predicted trajectories overlaid with ground truth paths, highlighting the effect of applying thresholds at 90\%, 80\%, and 70\%. 
In these plots, the reduction of noisy predictions and the improved alignment with the ground truth after filtering are clearly observable.

Table \ref{tab:uncertainty_combined} illustrates the changes in position and orientation errors under varying uncertainty thresholds applied to the TheGardenParty, ETRI, SusungHotel datasets.
The results demonstrate consistent reductions in mean position and orientation errors across all rejection thresholds (90\%, 80\%, and 70\%).
In the TheGardenParty dataset, applying a 70\% uncertainty threshold led to a reduction in the mean position error by as much as 12.1\% (from 0.776 m to 0.682 m)  and a decrease in the mean orientation error by up to 36.0\% (from 4.880$^{\circ}$ to 3.126$^{\circ}$)..
For the ETRI dataset under the same conditions, the reductions were 0.6
 and 29.2\% for orientation error (from 1.990$^{\circ}$ to 1.409$^{\circ}$).
Notably, the effectiveness of the uncertainty-based rejection strategy was also observed in the SusungHotel datasets.
For the SusungHotel dataset, the mean position error decreased from 4.262 m to 1.575 m, while the mean orientation error was reduced from 13.711$^{\circ}$ to 4.329$^{\circ}$, reflecting improvements of approximately 63.0
These findings indicate that the TheGardenParty and SusungHotel datasets have more outliers and noise, which leads to more significant performance improvements with the uncertainty-based rejection strategy.
In contrast, the ETRI dataset, likely collected in a more stable environment, shows lower errors even without applying the rejection method, resulting in relatively less impact from this approach.
Significantly, in the TheGardenParty dataset, the maximum position error decreased from 5.559 m to 4.192 m, and the maximum orientation error significantly dropped from 72.738$^{\circ}$ to 48.397$^{\circ}$.
This demonstrates the effectiveness of the rejection method in addressing extreme localization errors.
These results indicate our end-to-end localization with the uncertainty-based rejection method can be utilized as a global localization solution, which can also provide a reliable initial pose to conventional localization modules, e.g., the adaptive Monte Carlo localization, which is known as AMCL.

\section{Conclusions} \label{sec:conclusions}

This study experimentally demonstrated that our uncertainty-based rejection can effectively enhance robot localization performance.
By applying different rejection thresholds (90\%, 80\%, and 70\%), we confirmed that discarding network outputs with high uncertainty reduces both positional and orientation errors, thereby improving the reliability of model evaluation.
Unlike conventional end-to-end localization methods that treat all evaluations equally, the proposed approach improves the reliability of network outputs by selectively rejecting those with high uncertainty.
This method can be applied to other localization techniques based on deep neural networks.
However, it is essential to determine the appropriate value for the rejection threshold.
Thus, dynamically adjusting the rejection threshold based on dataset characteristics is necessary.
Future research will focus on optimizing the uncertainty-based rejection method for real-time robotic applications and evaluating its practicality through experiments on actual robots.
This approach will allow us to verify the effectiveness of the proposed rejection method in real-world environments and further enhance the reliability and accuracy of robot localization systems.

\backmatter





\bmhead{Acknowledgements}

This work was supported by Electronic and Telecommunications Research Institute (ETRI) grant funded by the Korean government [25ZD1130, Development of ICT Convergence Technology for Daegu-Gyeongbuk Regional Industry (Robots)].
We also sepecially thank to Dr. Hakjun Lee for his assistance with dataset collection.





\bibliography{Localization} 

\end{document}